\documentclass[11pt,a4paper]{article}
\usepackage[hyperref]{emnlp2020}
\usepackage{times}
\usepackage{latexsym}

\usepackage{microtype}

\aclfinalcopy %

\usepackage{amsmath}
\usepackage{amssymb}
\usepackage{hyperref}
\usepackage{booktabs}
\usepackage{pgfplots}
\pgfplotsset{width=.9\columnwidth}
\usepackage{multirow}
\usepackage{caption}
\usepackage{subcaption}
\usepackage{tikz}
\usepackage{xspace}
\usepackage{xargs}
\usepackage[colorinlistoftodos,prependcaption,textsize=tiny]{todonotes}
\usepackage{multirow}
\usetikzlibrary{positioning}
\usepackage{pifont}
\newcommand{\cmark}{\text{\ding{51}}}
\newcommand{\xmark}{\text{\ding{55}}}
\usepackage{makecell}

\newcommand\ignore[1]{}

\def\model/{DPR}

\newcommandx{\sy}[2][1=]{\todo[linecolor=purple,backgroundcolor=purple!10,bordercolor=purple,#1]{Scott: #2}\xspace}
\newcommandx{\dc}[2][1=]{\todo[linecolor=blue,backgroundcolor=blue!10,bordercolor=blue,#1]{Danqi: #2}\xspace}

\newcommand\ti[1]{\textit{#1}}
\newcommand\tf[1]{\textbf{#1}}
\newcommand\ttt[1]{\texttt{#1}}
\newcommand\mf[1]{\mathbf{#1}}

\newcommand*\samethanks[1][\value{footnote}]{\footnotemark[#1]}

\pgfplotsset{compat=1.14}

\title{Dense Passage Retrieval for Open-Domain Question Answering}

\author{\makecell{Vladimir Karpukhin\thanks{\hspace{.06in}Equal contribution}, Barlas O\u{g}uz\samethanks, Sewon Min$^\dagger$, Patrick Lewis, \\ Ledell Wu, Sergey Edunov, Danqi Chen$^\ddagger$, Wen-tau Yih} \\
Facebook AI \quad\quad  $^\dagger$University of Washington \quad\quad  $^\ddagger$Princeton University \\
\texttt{\{vladk, barlaso, plewis, ledell, edunov, scottyih\}@fb.com} \\
\texttt{sewon@cs.washington.edu} \\
\texttt{danqic@cs.princeton.edu} \\
}

\ignore{

\author{First Author \\
  Affiliation / Address line 1 \\
  Affiliation / Address line 2 \\
  Affiliation / Address line 3 \\
  \texttt{email@domain} \\\And
  Second Author \\
  Affiliation / Address line 1 \\
  Affiliation / Address line 2 \\
  Affiliation / Address line 3 \\
  \texttt{email@domain} \\}

}

\date{}

\begin{document}
\maketitle

\begin{abstract}
Open-domain question answering relies on efficient passage retrieval to select candidate contexts, where traditional sparse vector space models, such as TF-IDF or BM25, are the de facto method.
In this work, we show that retrieval can be practically implemented using \emph{dense} representations alone, where embeddings are learned from a small number of questions and passages by a simple dual-encoder framework.
When evaluated on a wide range of open-domain QA datasets, our dense retriever outperforms a strong Lucene-BM25 system greatly by 9\%-19\% absolute in terms of top-20 passage retrieval accuracy,
and helps our end-to-end QA system establish new state-of-the-art on multiple open-domain QA benchmarks.\footnote{The code and trained models have been released at \href{https://github.com/facebookresearch/DPR}{https://github.com/facebookresearch/DPR}.}

\end{abstract}

\section{Introduction}

\label{sec:intro}

Open-domain question answering (QA)~\cite{voorhees1999trec} is a task that answers factoid questions using a large collection of documents.
While early QA systems are often complicated and consist of multiple components~(\citet{ferrucci2012introduction,moldovan2003performance}, \textit{inter alia}),
the advances of reading comprehension models suggest a much simplified two-stage framework: (1) a context \emph{retriever} first selects a small subset of passages where some of them contain the answer to the question, and then (2) a machine \emph{reader} can thoroughly examine the retrieved contexts and identify the correct answer~\cite{chen2017reading}.
Although reducing open-domain QA to machine reading is a very reasonable strategy, a huge performance degradation is often observed in practice\footnote{For instance,
the exact match score on SQuAD v1.1 drops from above 80\% to less than 40\%~\cite{yang2019end}.}, indicating the needs of improving retrieval.

Retrieval in open-domain QA %
is usually implemented using TF-IDF or BM25~\cite{robertson2009probabilistic}, which matches keywords efficiently with an inverted index and can be seen
as representing the question and context in high-dimensional, sparse vectors (with weighting).
Conversely, the \emph{dense}, latent semantic encoding is \emph{complementary} to sparse representations by design.
For example, synonyms or paraphrases
that consist of completely different tokens may still be mapped to vectors close to each other.
Consider the question \ti{``Who is the bad guy in lord of the rings?''}, which can be answered from the context
\ti{``Sala Baker is best known for portraying the villain Sauron in the Lord of the Rings trilogy.''}
A term-based system would have difficulty retrieving such a context, while a dense retrieval system would be able to better match \ti{``bad guy''} with \ti{``villain''}
and fetch the correct context.
Dense encodings are also \emph{learnable} by adjusting the embedding functions, which provides additional flexibility to have a task-specific representation. With special in-memory data structures and indexing schemes, retrieval can be done efficiently using maximum inner product search (MIPS) algorithms~(e.g.,~\citet{NIPS2014_5329, guo2016quantization}).

However, it is generally believed that learning a good dense vector representation needs a large number of labeled pairs of question and contexts.
Dense retrieval methods have thus never be shown to outperform TF-IDF/BM25 for open-domain QA before ORQA~\cite{lee2019latent}, which proposes a sophisticated inverse cloze task (ICT) objective, predicting the blocks that contain the masked sentence, for additional pretraining.
The question encoder and the reader model are then fine-tuned using pairs of questions and answers jointly. %
Although ORQA successfully demonstrates that dense retrieval can outperform BM25, setting new state-of-the-art results on multiple open-domain QA datasets,
it also suffers from two weaknesses.
First, ICT pretraining is computationally intensive and it is not completely clear that regular sentences are good surrogates of questions in the objective function.
Second, because the context encoder is not fine-tuned using pairs of questions and answers, the corresponding representations could be suboptimal.

In this paper, we address the question: can we train a better dense embedding model using only pairs of questions and passages (or answers), \emph{without} additional pretraining?
By leveraging the now standard BERT pretrained model~\cite{devlin2018bert} and a dual-encoder architecture~\cite{bromley1994signature}, we focus on developing the right training scheme using a relatively small number of question and passage pairs. %
Through a series of careful ablation studies, our final solution is surprisingly simple: the embedding is optimized for maximizing inner products of the question and relevant passage vectors, with an objective comparing all pairs of questions and passages in a batch.
Our \emph{Dense Passage Retriever} (DPR) is exceptionally strong. It not only outperforms BM25 by a large margin (65.2\% vs. 42.9\% in Top-5 accuracy), but also results in a substantial improvement on the end-to-end QA accuracy compared to ORQA (41.5\% vs. 33.3\%) in the open Natural Questions setting~\cite{lee2019latent,kwiatkowski2019natural}.

Our contributions are twofold.
First, we demonstrate that with the proper training setup, simply fine-tuning the question and passage encoders on existing question-passage pairs
is sufficient to greatly outperform BM25.
Our empirical results also suggest that additional pretraining may not be needed.
Second, we verify that, in the context of open-domain question answering, a higher retrieval precision indeed translates to a higher end-to-end QA accuracy.
By applying a modern reader model to the top retrieved passages, we achieve comparable or better results on multiple QA datasets in the open-retrieval setting,
compared to several, much complicated systems.

\section{Background}
\label{sec:background}

The problem of open-domain QA studied in this paper can be described as follows.  Given a factoid question, such as
``\emph{Who first voiced Meg on Family Guy?}" or ``\emph{Where was the 8th Dalai Lama born?}", a system is required
to answer it using a large corpus of diversified topics.
More specifically, we assume the extractive QA setting, in which the answer is restricted to a span appearing in one or more passages in the corpus.
Assume that our collection contains $D$ documents, $d_1, d_2, \cdots, d_D$.
We first split each of the documents into text passages of equal lengths as the basic retrieval units\footnote{The ideal size and boundary of a text passage are functions of both the retriever and reader. We also experimented with natural paragraphs in our preliminary trials and found that using fixed-length passages performs better in both retrieval and final QA accuracy, as observed by~\citet{wang2019multi}.}
and get $M$ total passages in our corpus $\mathcal{C} = \{p_1, p_2, \ldots, p_{M}\}$, where
each passage $p_i$ can be viewed as a sequence of tokens $w^{(i)}_1, w^{(i)}_2, \cdots, w^{(i)}_{|p_i|}$.
Given a question $q$, the task is to find a span $w^{(i)}_s, w^{(i)}_{s+1}, \cdots, w^{(i)}_{e}$
from one of the passages $p_i$ that can answer the question.
Notice that %
to cover a wide variety of domains, the corpus size %
can easily range from millions of documents (e.g., Wikipedia) to billions (e.g., the Web).
As a result, any open-domain QA system needs to include an efficient \emph{retriever} component that
can select a small set of relevant texts, before applying the reader to extract the answer~\cite{chen2017reading}.\footnote{Exceptions include~\cite{seo2019real} and~\cite{roberts2020much}, which \emph{retrieves} and \emph{generates} the answers, respectively.}
Formally speaking, a retriever $R:(q,\mathcal{C}) \rightarrow \mathcal{C_F}$ is a function that takes as input a question $q$ and a corpus $\mathcal{C}$ and returns
a much smaller \emph{filter set} of texts $\mathcal{C_F} \subset \mathcal{C}$, where $|\mathcal{C_F}| = k \ll |\mathcal{C}|$.  For a fixed $k$, a \emph{retriever} can be evaluated in isolation on \emph{top-k retrieval accuracy}, which is the fraction of questions for which $\mathcal{C_F}$ contains a span that answers the question.

\section{Dense Passage Retriever (\model/)}
\label{sec:retriever}

We focus our research in this work on improving the \emph{retrieval} component in open-domain QA.
Given a collection of $M$ text passages, the goal of our dense passage retriever (\model/) is to index all the passages in a low-dimensional and continuous space,
such that it can retrieve efficiently the top $k$ passages relevant to the input question for the reader at run-time.
Note that $M$ can be very large
(e.g., 21 \text{million} passages in our experiments, described in Section~\ref{sec:wiki-preprocessing})
and $k$ is usually small, such as $20$--$100$.

\subsection{Overview}
Our dense passage retriever (\model/) uses a dense encoder $E_P(\cdot)$ which maps any text passage to a $d$-dimensional real-valued vectors and builds an index for all the $M$ passages that we will use for retrieval.
At run-time, \model/ applies a different encoder $E_Q(\cdot)$ that maps the input question to a $d$-dimensional vector, and retrieves $k$ passages
of which vectors are the closest to the question vector. We define the similarity between the question and the passage using the dot product of their vectors:
\begin{align}
    \mathrm{sim}(q, p) = E_Q(q)^{\intercal} E_P(p).
    \label{eq:sim}
\end{align}
Although more expressive model forms for measuring the similarity between a question and a passage do exist, such as networks consisting of multiple layers of cross attentions, the similarity function needs to be decomposable so that the representations of the collection of passages can be pre-computed.
Most decomposable similarity functions are some transformations of Euclidean distance (L2). For instance, cosine is equivalent to inner product for unit vectors and the Mahalanobis distance is equivalent to L2 distance in a transformed space.
Inner product search has been widely used and studied, as well as its connection to cosine similarity and L2 distance~\cite{mussmann2016learning,ram2012maximum}.
As our ablation study finds other similarity functions perform comparably (Section~\ref{sec:sim_loss}; Appendix~\ref{sec:alt-sim}), we thus choose the simpler inner product function and improve the dense passage retriever by learning better encoders.

\paragraph{Encoders}
Although in principle the question and passage encoders can be implemented by any neural networks, in this work we use two independent BERT~\cite{devlin2018bert} networks (base, uncased) and take the representation at the \ttt{[CLS]} token as the output, so $d = 768$.

\paragraph{Inference}
During inference time, we apply the passage encoder $E_P$ to all the passages and index them using FAISS \cite{johnson2017billion} offline.
FAISS is an extremely efficient, open-source library for similarity search and clustering of dense vectors, which can easily be applied to billions of vectors.
Given a question $q$ at run-time, we derive its embedding $v_q = E_Q(q)$ and retrieve the top $k$ passages with embeddings closest to $v_q$.

\subsection{Training}\label{sec:training}

Training the encoders so that the dot-product similarity (Eq.~\eqref{eq:sim}) becomes a good ranking function for retrieval is essentially a \emph{metric learning} problem~\cite{kulis2013metric}.
The goal is to create a vector space such that relevant pairs of questions and passages will have smaller distance (i.e., higher similarity) than the irrelevant ones, by learning a better embedding function.

Let $\mathcal{D} = \{ \langle q_i, p^+_i, p^-_{i,1}, \cdots, p^-_{i,n} \rangle \}_{i=1}^m$ be the training data that consists of $m$ instances.
Each instance contains one question $q_i$ and one relevant (positive) passage $p^+_i$, along with $n$ irrelevant (negative) passages $p^-_{i,j}$.
We optimize the loss function as the negative log likelihood of the positive passage:
\begin{eqnarray}
&& L(q_i, p^+_i, p^-_{i,1}, \cdots, p^-_{i,n}) \label{eq:training} \\
&=& -\log \frac{ e^{\mathrm{sim}(q_i, p_i^+)} }{e^{\mathrm{sim}(q_i, p_i^+)} + \sum_{j=1}^n{e^{\mathrm{sim}(q_i, p^-_{i,j})}}}. \nonumber
\end{eqnarray}

\paragraph{Positive and negative passages}
For retrieval problems, it is often the case that positive examples are available explicitly, while negative examples need to be selected from an extremely large pool.
For instance, passages relevant to a question may be given in a QA dataset, or can be found using the answer.
All other passages in the collection, while not specified explicitly, can be viewed as irrelevant by default.
In practice, how to select negative examples is often overlooked but could be decisive for learning a high-quality encoder.
We consider three different types of negatives: (1) Random: any random passage from the corpus; (2) BM25: top passages returned by BM25 which don't contain the answer but match most question tokens; (3) Gold: positive passages paired with other questions which appear in the training set. We will discuss the impact of different types of negative passages and training schemes in Section~\ref{sec:ir_ablation}.
Our best model uses gold passages from the same mini-batch and one BM25 negative passage.
In particular, re-using gold passages from the same batch as negatives can make the computation efficient while achieving great performance.
We discuss this approach below.

\paragraph{In-batch negatives}
Assume that we have $B$ questions in a mini-batch and each one is associated with a relevant passage. Let $\mathbf{Q}$ and $\mathbf{P}$ be the $(B\times d)$ matrix of question and passage embeddings in a batch of size $B$.
$\mathbf{S} = \mathbf{Q}\mathbf{P}^T$ is a $(B\times B)$ matrix of similarity scores, where each row of which corresponds to a question, paired with $B$ passages.
In this way, we reuse computation and effectively train on $B^2$ ($q_i$, $p_j$) question/passage pairs in each batch.
Any ($q_i$, $p_j$) pair is a positive example when $i=j$, and negative otherwise.
This creates $B$ training instances in each batch, where there are $B-1$ negative passages for each question.

The trick of in-batch negatives has been used in the full batch setting~\cite{yih2011learning} and more recently
for mini-batch~\cite{henderson2017efficient,gillick-etal-2019-learning}.
It has been shown to be an effective strategy for learning a dual-encoder model %
that boosts the number of training examples.

\section{Experimental Setup}
\label{sec:exp-setup}

In this section, we describe the data we used for experiments and the basic setup.

\subsection{Wikipedia Data Pre-processing}
\label{sec:wiki-preprocessing}

Following~\cite{lee2019latent}, we use the English Wikipedia dump from Dec.~20, 2018 as the source documents for answering questions.
We first apply the pre-processing code released in DrQA~\cite{chen2017reading} to extract the clean, text-portion of articles from the Wikipedia dump.
This step removes semi-structured data, such as tables, info-boxes, lists, as well as the disambiguation pages.
We then split each article into multiple, disjoint text blocks of 100 words as \emph{passages}, serving as our basic retrieval units, following~\cite{wang2019multi}, which results in 21,015,324 passages in the end.\footnote{However, \citet{wang2019multi} also propose splitting documents into overlapping passages, which we do not find advantageous compared to the non-overlapping version.}
Each passage is also prepended with the title of the Wikipedia article where the passage is from, along with an \ttt{[SEP]} token.%

\subsection{Question Answering Datasets}
\label{sec:qa-datasets}

We use the same five QA datasets and training/dev/testing splitting method as in previous work~\cite{lee2019latent}.
Below we briefly describe each dataset and refer readers to their paper for the details of data preparation.

\noindent
\textbf{Natural Questions (NQ)}~\cite{kwiatkowski2019natural} was designed for end-to-end question answering.
The questions were mined from real Google search queries and the answers were spans in Wikipedia articles identified by annotators.

\noindent
\textbf{TriviaQA}~\cite{joshi-etal-2017-triviaqa} contains a set of trivia questions with answers that were originally scraped from the Web.

\noindent
\textbf{WebQuestions (WQ)}~\cite{berant2013semantic} consists of questions selected using Google Suggest API, where the answers are entities in Freebase.

\noindent
\textbf{CuratedTREC (TREC)}~\cite{baudivs2015modeling} sources questions from TREC QA tracks as well as various Web sources
and is intended for open-domain QA from unstructured corpora.

\noindent
\textbf{SQuAD v1.1}~\cite{rajpurkar2016squad} is a popular benchmark dataset for reading comprehension.
Annotators were presented with a Wikipedia paragraph, and asked to write questions that could be answered from the given text.
Although SQuAD has been used previously for open-domain QA research, it is not ideal because many questions lack context in absence of the provided paragraph.
We still include it in our experiments for providing a fair comparison to previous work and we will discuss more in Section~\ref{sec:retrieval_main_results}.

\paragraph{Selection of positive passages} %

Because only pairs of questions and answers are provided in TREC, WebQuestions and TriviaQA\footnote{We use the unfiltered TriviaQA version and discard the noisy evidence documents mined from Bing.}, we use the highest-ranked passage from BM25 that contains the answer as the positive passage.  If none of the top 100 retrieved passages has the answer, the question will be discarded.
For SQuAD and Natural Questions, since the original passages have been split and processed differently than our pool of candidate passages, we match and replace each gold passage with the corresponding passage in the candidate pool.\footnote{The improvement of using gold contexts over passages that contain answers is small. See Section~\ref{sec:pos_pass} and Appendix~\ref{sec:distant}.}
We discard the questions when the matching is failed due to different Wikipedia versions or pre-processing.
Table~\ref{tab:split-stats} shows the number of questions in training/dev/test sets for all the datasets and the actual questions used for training the retriever.

\begin{table}[!t]
    \small
    \centering
    \begin{tabular}{lrrrr} \toprule
    \tf{Dataset} & \multicolumn{2}{c}{\tf{Train}}  & \tf{Dev} & \tf{Test} \\
    \midrule
    Natural Questions   & 79,168 & 58,880 & 8,757 & 3,610 \\
    TriviaQA            & 78,785 & 60,413 & 8,837 & 11,313  \\
    WebQuestions        & 3,417  & 2,474 & 361   & 2,032 \\
    CuratedTREC         & 1,353  & 1,125 & 133   & 694 \\
    SQuAD               & 78,713 & 70,096 & 8,886 & 10,570  \\
    \bottomrule
    \end{tabular}
    \caption{Number of questions in each QA dataset. The two columns of \textbf{Train} denote the original training examples in the dataset and the actual questions used for training DPR after filtering.  See text for more details.}
    \label{tab:split-stats}
\end{table}

\ignore{
\begin{table*}[!ht]
    \centering
    \begin{tabular}{llcccc} \toprule
    $k$ & Retriever & {Natural Questions} & {TriviaQA} & {WebQuestions} & {CuratedTREC} \\
    \midrule
    \multirow{2}{*}{5} &
    BM25   & 41.6 & 54.4 & 36.8 & 54.9  \\
    & Dense  & \tf{67.0} & \tf{69.7} & \tf{63.9} & \tf{79.9}  \\
    & Dense+ & & & & \\
    \midrule
    \multirow{2}{*}{20} &
        BM25   & 57.9 & 66.9 & 55.0 & 70.9  \\
    & Dense  & \tf{79.4} & \tf{78.8} & \tf{75.0} & \tf{89.1}  \\
    & Dense+ & & & & \\
    \midrule
    \multirow{2}{*}{50} &
        BM25   & 67.7 & 72.8 & 64.5 & 80.3  \\
    & Dense  & \tf{83.8} & \tf{82.5} & \tf{80.5} & \tf{92.6}  \\
    & Dense+ & & & & \\
    \midrule
    \multirow{2}{*}{100} &
        BM25   & 73.2 & 76.7 & 71.1 & 84.1 \\
    & Dense  & \tf{86.0} & \tf{84.7} & \tf{82.9} & \tf{93.9}  \\
    & Dense+ & & & & \\
    \bottomrule
    \end{tabular}
    \caption{Top-$K$ retrieval accuracy, measured as the percentage that top $k$ retrieved passages that contain an answer. The numbers are reported on the testing set.}
    \label{tab:qa_ir}
\end{table*}

}

\begin{table*}[t!]
    \setlength\tabcolsep{5pt}
    \centering
    \small
    \begin{tabular}{ll|ccccc|ccccc}
    \toprule
    \tf{Training} & \tf{Retriever} & \multicolumn{5}{c}{\tf{Top-20}} & \multicolumn{5}{|c}{\tf{Top-100}} \\
    & & NQ & TriviaQA & WQ & TREC & SQuAD & NQ & TriviaQA & WQ & TREC & SQuAD \\ \midrule
    None & BM25 & 59.1 & 66.9 & 55.0 & 70.9 & 68.8 & 73.7 & 76.7 & 71.1 & 84.1 & 80.0 \\
    \midrule
    \multirow{2}{*}{Single} &\model/ & 78.4 & 79.4 & 73.2 & 79.8 & 63.2 & 85.4 & \textbf{85.0} & 81.4 & 89.1 & 77.2 \\
    &BM25 + \model/ & 76.6 & 79.8 & 71.0 & 85.2 & \textbf{71.5} & 83.8 & 84.5 & 80.5 & 92.7 & \textbf{81.3} \\
    \midrule
    \multirow{2}{*}{Multi} &\model/ & \textbf{79.4} & 78.8 & \textbf{75.0} & \textbf{89.1} & 51.6 & \textbf{86.0} & 84.7 & \textbf{82.9} & 93.9 & 67.6 \\
    &BM25 + \model/ & 78.0 & \textbf{79.9} & 74.7 & 88.5 & 66.2 & 83.9 & 84.4 & 82.3 & \textbf{94.1} & 78.6 \\ \bottomrule
    \end{tabular}
     \caption{Top-20 \& Top-100 retrieval accuracy on test sets, measured as the percentage of top 20/100 retrieved passages that contain the answer. \ti{Single} and \ti{Multi} denote that our Dense Passage Retriever (DPR) was trained using individial or combined training datasets (all the datasets excluding SQuAD). See text for more details.} %

    \label{tab:qa_ir}
\end{table*}

\section{Experiments: Passage Retrieval}
\label{sec:exp-psg}

In this section, we evaluate the retrieval performance of our Dense Passage Retriever (\model/), along with analysis on
how its output differs from traditional retrieval methods, the effects of different training schemes and the run-time efficiency.

The \model/ model used in our main experiments is trained using the in-batch negative setting (Section~\ref{sec:training}) with a batch size of $128$ and one additional BM25 negative passage per question.
We trained the question and passage encoders for up to $40$ epochs for large datasets (NQ, TriviaQA, SQuAD) and $100$ epochs for small datasets (TREC, WQ), with a learning rate of $10^{-5}$ using Adam, linear scheduling with warm-up and dropout rate~$0.1$.

While it is good to have the flexibility to adapt the retriever to each dataset, it would also be desirable to obtain a single retriever that works well across the board. To this end, we train a \emph{multi}-dataset encoder by combining training data from all datasets excluding SQuAD.\footnote{SQuAD is limited to a small set of Wikipedia documents and thus introduces unwanted bias. We will discuss this issue more in Section~\ref{sec:retrieval_main_results}.} %
In addition to \model/, we also present the results of BM25, the traditional retrieval method\footnote{\href{https://lucene.apache.org/}{Lucene} implementation. BM25 parameters $b=0.4$ (document length normalization) and $k_1=0.9$ (term frequency scaling) are tuned using development sets.} and BM25+\model/, using a linear combination of their scores as the new ranking function. 
Specifically, we obtain two initial sets of top-2000 passages based on BM25 and DPR, respectively, and rerank the union of them using BM25($q$,$p$) + $\lambda \cdot \mathrm{sim}(q,p)$ as the ranking function. We used $\lambda=1.1$ based on the retrieval accuracy in the development set.

\subsection{Main Results}
\label{sec:retrieval_main_results}

Table~\ref{tab:qa_ir} compares different passage retrieval systems on five QA datasets, using the top-$k$ accuracy ($k \in \{20,100\}$).
With the exception of SQuAD, \model/ performs consistently better than BM25 on all datasets.  The gap is especially large when $k$ is small (e.g., 78.4\% vs. 59.1\% for top-20 accuracy on Natural Questions).
When training with multiple datasets, TREC, the smallest dataset of the five, benefits greatly from more training examples.  In contrast, Natural Questions and WebQuestions improve modestly and TriviaQA degrades slightly.
Results can be improved further in some cases by combining \model/ with BM25 in both single- and multi-dataset settings.

We conjecture that the lower performance on SQuAD is due to two reasons. First, the annotators wrote questions after seeing the passage.  As a result, there is a high lexical overlap between passages and questions, which gives BM25 a clear advantage.  Second, the data was collected from only 500+ Wikipedia articles and thus
the distribution of training examples is extremely biased, as argued previously by~\newcite{lee2019latent}.

\begin{figure}[!t]
    \center
    \includegraphics[scale=0.4,trim=0 0 0 40, clip]{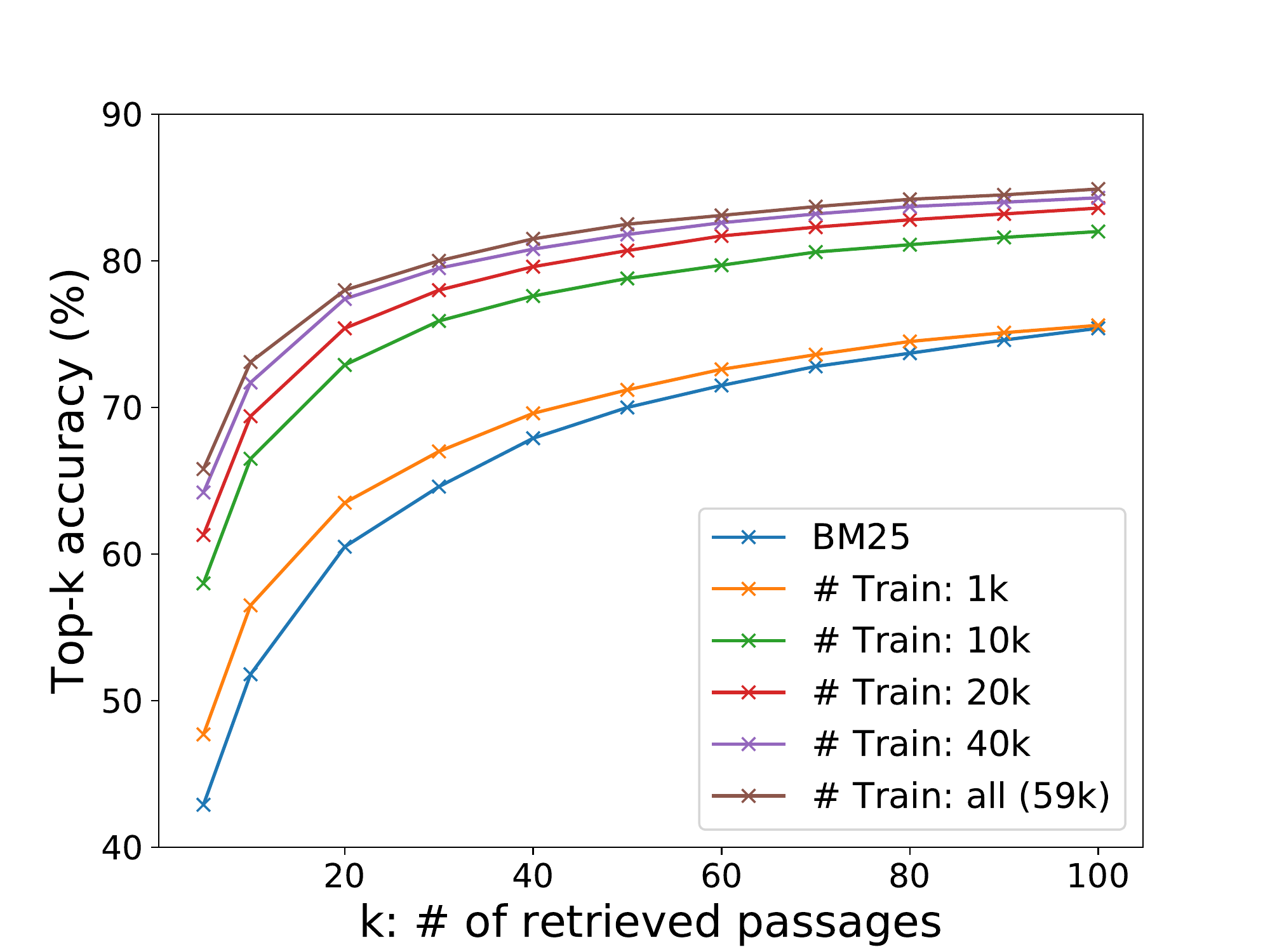}
    \caption{Retriever top-$k$ accuracy with different numbers of training examples used in our dense passage retriever vs BM25. The results are measured on the development set of Natural Questions. Our DPR trained using 1,000 examples already outperforms BM25. }
    \label{fig:ir-training-examples}
\end{figure}

\subsection{Ablation Study on Model Training}
\label{subsec:ablation-retrieval}

To understand further how different model training options affect the results, we conduct several additional experiments and discuss our findings below.

\paragraph{Sample efficiency}

We explore how many training examples are needed to achieve good passage retrieval performance. %
Figure~\ref{fig:ir-training-examples} illustrates the top-$k$ retrieval accuracy with respect to different numbers of training examples,
measured on the development set of Natural Questions.
As is shown, a dense passage retriever trained using only 1,000 examples already outperforms BM25. 
This suggests that with a general pretrained language model, it is possible to train a high-quality dense retriever with a small number of question--passage pairs.
Adding more training examples (from 1k to 59k) further improves the retrieval accuracy consistently.

\paragraph{In-batch negative training}
\label{sec:ir_ablation}
We test different training schemes on the development set of Natural Questions and summarize the results in Table~\ref{tab:ir-ablation}. 
The top block is the standard 1-of-$N$ training setting, where each question in the batch is paired with a positive passage and its own set of $n$ negative passages~(Eq.~\eqref{eq:training}).
We find that the choice of negatives --- random, BM25 or gold passages (positive passages from other questions) --- does not impact the top-$k$ accuracy much in this setting when $k \ge 20$.%

The middle bock is the in-batch negative training (Section~\ref{sec:training}) setting.
We find that using a similar configuration (7 gold negative passages), in-batch negative training improves the results substantially.
The key difference between the two is whether the gold negative passages come from the same batch or from the whole training set.
Effectively, in-batch negative training is an easy and memory-efficient way to reuse the negative examples already in the batch rather than creating new ones.
It produces more pairs and thus increases the number of training examples, which might contribute to the good model performance. 
As a result, accuracy consistently improves as the batch size grows.

Finally, we explore in-batch negative training with additional ``hard" negative passages that have high BM25 scores given the question, but do not contain the answer string (the bottom block). These additional passages are used as negative passages for all questions in the same batch.
We find that adding a single BM25 negative passage improves the result substantially while adding two does not help further.

\begin{table}[t]
    \centering
    \small
    \begingroup
    \setlength{\tabcolsep}{0.70\tabcolsep}
    \begin{tabular}{@{}l l c cc c c @{}} \toprule
    \tf{Type} & \tf{\#N} & \tf{IB} & \tf{Top-5} & \tf{Top-20} & \tf{Top-100} \\ \midrule
    Random & 7 & $\xmark$ & 47.0 & 64.3 & 77.8 \\
    BM25 & 7 & $\xmark$ & 50.0 & 63.3 & 74.8\\
    Gold & 7 & $\xmark$ & 42.6 & 63.1 & 78.3\\
    \midrule
    Gold & 7 & $\cmark$ & 51.1 & 69.1 & 80.8 \\
    Gold & 31 & $\cmark$ & 52.1 & 70.8 & 82.1 \\
    Gold & 127 & $\cmark$ & 55.8 & 73.0 & 83.1\\
    \midrule
    G.+BM25$^{(1)}$ & 31+32 & $\cmark$ & 65.0 & 77.3 & 84.4 \\
    {G.+BM25}$^{(2)}$ & 31+64 & $\cmark$ & 64.5 & 76.4 & 84.0 \\
    G.+BM25$^{(1)}$ & 127+128 & $\cmark$ & \tf{65.8} & \tf{78.0} & \tf{84.9} \\
    \bottomrule
    \end{tabular}
    \endgroup
    \caption{Comparison of different training schemes, measured as top-$k$ retrieval accuracy on Natural Questions (development set).  \#N: number of negative examples, IB: in-batch training. G.+BM25$^{(1)}$ and {G.+BM25}$^{(2)}$ denote in-batch training with 1 or 2 additional BM25 negatives, which serve as negative passages for all questions in the batch.}

    \label{tab:ir-ablation}
\end{table}

\paragraph{Impact of gold passages} %
\label{sec:pos_pass}
We use passages that match the gold contexts in the original datasets (when available) as positive examples (Section~\ref{sec:qa-datasets}).
Our experiments on Natural Questions show that switching to distantly-supervised passages (using the highest-ranked BM25 passage that contains the answer), has only a small impact: 1 point lower top-$k$ accuracy for retrieval. %
Appendix~\ref{sec:distant} contains more details.

\paragraph{Similarity and loss}
\label{sec:sim_loss}
Besides dot product, cosine and Euclidean L2 distance are also commonly used as decomposable similarity functions.
We test these alternatives and find that L2 performs comparable to dot product, and both of them are superior to cosine.
Similarly, in addition to negative log-likelihood, a popular option for ranking is triplet loss, which compares a positive passage and a negative one directly with respect to a question~\cite{burges2005learning}.
Our experiments show that using triplet loss does not affect the results much.
More details can be found in Appendix~\ref{sec:alt-sim}. %

\paragraph{Cross-dataset generalization}
One interesting question regarding \model/'s discriminative training is how much performance degradation it may suffer from a non-iid setting. In other words, can it still generalize well when directly applied to a different dataset without additional fine-tuning? To test the cross-dataset generalization, we train \model/ on Natural Questions only and test it directly on the smaller WebQuestions and CuratedTREC datasets. We find that \model/ generalizes well, with 3-5 points loss from the best performing fine-tuned model in top-20 retrieval accuracy (69.9/86.3 vs. 75.0/89.1 for WebQuestions and TREC, respectively), while still greatly outperforming the BM25 baseline (55.0/70.9).

\subsection{Qualitative Analysis}

Although \model/ performs better than BM25 in general, passages retrieved by these two methods differ qualitatively. 
Term-matching methods like BM25 are sensitive to highly selective keywords and phrases, while \model/ captures lexical variations or semantic relationships better.
See Appendix~\ref{sec:qual_ana} for examples and more discussion.

\subsection{Run-time Efficiency}
\label{sec:efficiency}

The main reason that we require a retrieval component for open-domain QA is to reduce the number of candidate passages that the reader needs to consider, which is crucial for answering user's questions in real-time.
We profiled the passage retrieval speed on a server with Intel Xeon CPU E5-2698 v4 @ 2.20GHz and 512GB memory.
With the help of FAISS in-memory index for real-valued vectors\footnote{FAISS configuration: we used HNSW index type on CPU, neighbors to store per node = 512, construction time search depth = 200,  search depth = 128.}, \model/ can be made incredibly efficient, processing 995.0 questions per second, returning top 100 passages per question.
In contrast, BM25/Lucene (implemented in Java, using file index) processes 23.7 questions per second per CPU thread.

On the other hand, the time required for building an index for dense vectors is much longer.
Computing dense embeddings on 21-million passages is resource intensive, but can be easily parallelized, taking roughly 8.8 hours on 8 GPUs.
However, building the FAISS index on 21-million vectors on a single server takes 8.5 hours.
In comparison, building an inverted index using Lucene is much cheaper and takes only about 30 minutes in total.

\newcommand{\nqbm}{32.6}
\newcommand{\nqsingle}{\tf{41.5}}
\newcommand{\nqsinglehybrid}{39.0}
\newcommand{\nqmulti}{\tf{41.5}}
\newcommand{\nqmultihybrid}{38.8}

\newcommand{\triviabm}{52.4}
\newcommand{\triviasingle}{56.8}
\newcommand{\triviasinglehybrid}{57.0}
\newcommand{\triviamulti}{56.8}
\newcommand{\triviamultihybrid}{\tf{57.9}}

\newcommand{\sqbm}{38.1}
\newcommand{\sqsingle}{29.8}
\newcommand{\sqsinglehybrid}{36.7}
\newcommand{\sqmulti}{24.1}
\newcommand{\sqmultihybrid}{35.8}

\newcommand{\webqbm}{29.9}
\newcommand{\webqsingle}{34.6}
\newcommand{\webqsinglehybrid}{35.2}
\newcommand{\webqmulti}{\tf{42.4}}
\newcommand{\webqmultihybrid}{41.1}

\newcommand{\trecbm}{24.9}
\newcommand{\trecsingle}{25.9}
\newcommand{\trecsinglehybrid}{28.0}
\newcommand{\trecmulti}{49.4}
\newcommand{\trecmultihybrid}{\tf{50.6}}

\begin{table*}[t]
    \setlength\tabcolsep{5pt}
    \centering
    \begin{tabular}{llccccc} \toprule
    \tf{Training} & \tf{Model} & \tf{NQ} & \tf{TriviaQA} & \tf{WQ} & \tf{TREC} & \tf{SQuAD} \\ \midrule
    Single & {BM25+BERT}~\cite{lee2019latent} & 26.5 & 47.1 & 17.7 & 21.3 & 33.2 \\
    Single & {ORQA}~\cite{lee2019latent}  & 33.3 & 45.0 & 36.4 & 30.1 & 20.2   \\
    Single & {HardEM}~\cite{min2019discrete} & 28.1 & 50.9 & - & - & - \\
    Single & {GraphRetriever}~\cite{min2019knowledge} & 34.5 & 56.0 & 36.4 & - & - \\
    Single & {PathRetriever}~\cite{asai2020learning}  & 32.6 & - & - & - & \tf{56.5}\\
    Single & {REALM}$_\textrm{Wiki}$~\cite{guu2020realm} & 39.2 & - & 40.2 & 46.8 & - \\ 
    Single & {REALM}$_\textrm{News}$~\cite{guu2020realm} & 40.4 & - & 40.7 & 42.9 & - \\ 
    \midrule
    \multirow{3}{*}{Single} & BM25   & \nqbm & \triviabm & \webqbm & \trecbm & \sqbm \\
    &\model/  & \nqsingle & \triviasingle & \webqsingle & \trecsingle & \sqsingle \\
    &BM25+\model/  & \nqsinglehybrid & \triviasinglehybrid & \webqsinglehybrid & \trecsinglehybrid & \sqsinglehybrid \\
    \midrule
    \multirow{2}{*}{Multi} & \model/  & \nqmulti & \triviamulti & \webqmulti & \trecmulti & \sqmulti \\
    &BM25+\model/ & \nqmultihybrid & \triviamultihybrid & \webqmultihybrid & \trecmultihybrid & \sqmultihybrid \\
    \bottomrule
    \end{tabular}
    \caption{End-to-end QA (Exact Match) Accuracy. The first block of results are copied from their cited papers. {REALM}$_\textrm{Wiki}$ and {REALM}$_\textrm{News}$ are the same model but pretrained on Wikipedia and CC-News, respectively.
    \ti{Single} and \ti{Multi} denote that our Dense Passage Retriever (DPR) is trained using individual or combined training datasets (all except SQuAD). For WQ and TREC in the \ti{Multi} setting, we fine-tune the reader trained on NQ.}
    \label{tab:qa_em}
\end{table*}

\section{Experiments: Question Answering}
\label{sec:exp-qa}

In this section, we experiment with how different passage retrievers affect the final QA accuracy.

\subsection{End-to-end QA System}

We implement an end-to-end question answering system in which we can plug different retriever systems directly. 
Besides the retriever, our QA system consists of a neural \textit{reader} that outputs the answer to the question.
Given the top $k$ retrieved passages (up to $100$ in our experiments), the reader assigns a passage selection score to each passage.  
In addition, it extracts an answer span from each passage and assigns a span score.
The best span from the passage with the highest passage selection score is chosen as the final answer.
The passage selection model serves as a reranker through cross-attention between the question and the passage. 
Although cross-attention is not feasible for retrieving relevant passages in a large corpus due to its non-decomposable nature, it has more capacity than the dual-encoder model $\mathrm{sim}(q,p)$ as in Eq.~\eqref{eq:sim}. Applying it to selecting the passage from a small number of retrieved candidates has been shown to work well~\cite{wang2019multi, wang2018r, lin2018denoising}.

Specifically, let $\mathbf{P}_{i} \in \mathbb{R}^{L \times h}$ ($1 \leq i \leq k$) be a BERT (base, uncased in our experiments) representation for the $i$-th passage, where $L$ is the maximum length of the passage and $h$ the hidden dimension.
The probabilities of a token being the starting/ending positions of an answer span and a passage being selected are defined as:
\begin{eqnarray}
    P_{\textrm{start},i}(s)  &=& \mathrm{softmax} \big( \mathbf{P}_{i} \mf{w}_\textrm{start}\big)_s, \\
    P_{\textrm{end},i}(t) &=& \mathrm{softmax} \big( \mathbf{P}_{i} \mf{w}_\textrm{end}\big)_t, \\
    P_\textrm{selected}(i) &=& \mathrm{softmax} \big( \mathbf{\hat{P}}^{\intercal} \mf{w}_\mathrm{selected}  \big)_i,
\end{eqnarray} where $\mathbf{\hat{P}} = [\mathbf{P}_1^{\mathrm{[CLS]}}, \ldots, \mathbf{P}_k^{\mathrm{[CLS]}}] \in \mathbb{R}^{h \times k}$ and $\mf{w}_\textrm{start}, \mf{w}_\textrm{end},\mf{w}_\mathrm{selected} \in \mathbb{R}^{h}$ are learnable vectors. 
We compute a span score of the $s$-th to $t$-th words from the $i$-th passage as $P_{\textrm{start},i}(s) \times P_{\textrm{end},i}(t)$, and a passage selection score of the $i$-th passage as $P_\textrm{selected}(i)$.

During training, we sample one positive and $\tilde{m}-1$ negative passages from the top 100 passages returned by the retrieval system (BM25 or DPR) for each question.  $\tilde{m}$ is a hyper-parameter and we use $\tilde{m}=24$ in all the experiments. The training objective is to maximize the marginal log-likelihood of all the correct answer spans in the positive passage (the answer string may appear multiple times in one passage), combined with the log-likelihood of the positive passage being selected. We use the batch size of 16 for large (NQ, TriviaQA, SQuAD) and 4 for small (TREC, WQ) datasets, and tune $k$ on the development set. For experiments on small datasets under the {\em Multi} setting, in which using other datasets is allowed, we fine-tune the reader trained on Natural Questions to the target dataset. All experiments were done on eight 32GB GPUs.

\subsection{Results}

Table~\ref{tab:qa_em} summarizes our final end-to-end QA results, measured by \emph{exact match} with the reference answer after minor normalization as  in~\cite{chen2017reading,lee2019latent}.  From the table, we can see that higher retriever accuracy typically leads to better final QA results: 
in all cases except SQuAD, answers extracted from the passages retrieved by \model/ are more likely to be correct, compared to those from BM25.
For large datasets like NQ and TriviaQA, models trained using multiple datasets (Multi) perform comparably to those trained using the individual training set (Single).  Conversely, on smaller datasets like WQ and TREC, the multi-dataset setting has a clear advantage. Overall, our \model/-based models outperform the previous state-of-the-art results on four out of the five datasets, with 1\% to 12\% absolute differences in exact match accuracy.
It is interesting to contrast our results to those of ORQA~\cite{lee2019latent} and also the concurrently developed approach, REALM~\cite{guu2020realm}. While both methods include additional pretraining tasks and employ an expensive end-to-end training regime, \model/ manages to outperform them on both NQ and TriviaQA, simply by focusing on learning a strong passage retrieval model using pairs of questions and answers.
The additional pretraining tasks are likely more useful only when the target training sets are small. Although the results of \model/ on WQ and TREC in the single-dataset setting are less competitive, adding more question--answer pairs helps boost the performance, achieving the new state of the art.

To compare our pipeline training approach with joint learning, we run an ablation on Natural Questions where the retriever and reader are jointly trained, following \citet{lee2019latent}.
This approach obtains a score of 39.8 EM, which suggests that our strategy of training a strong retriever and reader in isolation can leverage effectively available supervision, while outperforming a comparable joint training approach with a simpler design (Appendix~\ref{sec:joint}).

One thing worth noticing is that our reader does consider more passages compared to ORQA, although it is not completely clear how much more time it takes for inference. 
While \model/ processes up to 100 passages for each question, the reader is able to fit all of them into one batch on a single 32GB GPU, thus the latency remains almost identical to the single passage case (around 20ms).
The exact impact on throughput is harder to measure: ORQA uses 2-3x longer passages compared to DPR (288 word pieces compared to our 100 tokens) and the computational complexity is super-linear in passage length. We also note that we found $k=50$ to be optimal for NQ, and $k=10$ leads to only marginal loss in exact match accuracy (40.8 vs. 41.5 EM on NQ), which should be roughly comparable to ORQA's 5-passage setup.

\section{Related Work}
\label{sec:related}

Passage retrieval has been %
an important component for open-domain QA~\cite{voorhees1999trec}.
It not only effectively reduces the search space for answer extraction, but also identifies the support context for users to verify the answer.
Strong sparse vector space models like TF-IDF or BM25 have been used as the standard method applied broadly to various QA tasks~\citep[e.g.,][]{chen2017reading, yang2019end, Yang2019Data, nie2019revealing, min2019discrete, wolfson2020break}.
Augmenting text-based retrieval with external structured information, such as knowledge graph and Wikipedia hyperlinks, has also been explored recently~\cite{min2019knowledge,asai2020learning}.

The use of dense vector representations for retrieval has a long history since Latent Semantic Analysis~\cite{deerwester1990indexing}.
Using labeled pairs of queries and documents, discriminatively trained dense encoders have become popular recently~\cite{yih2011learning, huang2013learning, gillick-etal-2019-learning}, with applications to cross-lingual document retrieval, ad relevance prediction, Web search and entity retrieval. Such approaches complement the sparse vector methods as they can potentially give high similarity scores to semantically relevant text pairs, even without exact token matching. The dense representation alone, however, is typically inferior to the sparse one.
While not the focus of this work, dense representations from pretrained models, along with cross-attention mechanisms, have also been shown effective in passage or dialogue re-ranking tasks~\cite{nogueira2019passage,humeau2020poly}. 
Finally, a concurrent work~\cite{khattab2020colbert} demonstrates the feasibility of full dense retrieval in IR tasks. Instead of employing the dual-encoder framework, they introduced a late-interaction operator on top of the BERT encoders.

Dense retrieval for open-domain QA has been explored by~\citet{das2019multi}, who propose to retrieve relevant passages iteratively using reformulated question vectors.
As an alternative approach that skips passage retrieval, \citet{seo2019real} propose to encode candidate answer phrases as vectors and directly retrieve the answers to the input questions efficiently.
Using additional pretraining with the objective that matches surrogates of questions and relevant passages, \citet{lee2019latent} jointly train the question encoder and reader.
Their approach outperforms the BM25 plus reader paradigm on multiple open-domain QA datasets in QA accuracy, and is further extended by REALM~\cite{guu2020realm}, which includes tuning the passage encoder asynchronously by re-indexing the passages during training.
The pretraining objective has also recently been improved by \citet{Xiong2020ProgressivelyPD}.
In contrast, our model provides a simple and yet effective solution that shows stronger empirical performance, without relying on additional pretraining or complex joint training schemes.

\model/ has also been used as an important module in very recent work. For instance, extending the idea of leveraging hard negatives, \citet{Xiong2020ANCE} use the retrieval model trained in the previous iteration to discover new negatives and construct a different set of examples in each training iteration. Starting from our trained \model/ model, they show that the retrieval performance can be further improved. Recent work~\cite{Izacard2020fid, lewis2020rag} have also shown that DPR can be combined with generation models such as BART~\cite{lewis-etal-2020-bart} and T5~\cite{raffel2019T5}, achieving good performance on open-domain QA and other knowledge-intensive tasks.

\section{Conclusion}
\label{sec:conclusion}

In this work, we demonstrated that dense retrieval can outperform and potentially replace the traditional sparse retrieval component in open-domain question answering. While a simple dual-encoder approach can be made to work surprisingly well, we showed that there are some critical ingredients to training a dense retriever successfully. Moreover, our empirical analysis and ablation studies indicate that more complex model frameworks or similarity functions do not necessarily provide additional values. As a result of improved retrieval performance, we obtained new state-of-the-art results on multiple open-domain question answering benchmarks.

\section*{Acknowledgments}
We thank the anonymous reviewers for their helpful comments and suggestions. 

\bibliography{emnlp2020}

\begin{thebibliography}{48}
\expandafter\ifx\csname natexlab\endcsname\relax\def\natexlab#1{#1}\fi

\bibitem[{Asai et~al.(2020)Asai, Hashimoto, Hajishirzi, Socher, and
  Xiong}]{asai2020learning}
Akari Asai, Kazuma Hashimoto, Hannaneh Hajishirzi, Richard Socher, and Caiming
  Xiong. 2020.
\newblock Learning to retrieve reasoning paths over {Wikipedia} graph for
  question answering.
\newblock In \emph{International Conference on Learning Representations
  (ICLR)}.

\bibitem[{Baudi{\v{s}} and {\v{S}}ediv{\`y}(2015)}]{baudivs2015modeling}
Petr Baudi{\v{s}} and Jan {\v{S}}ediv{\`y}. 2015.
\newblock Modeling of the question answering task in the yodaqa system.
\newblock In \emph{International Conference of the Cross-Language Evaluation
  Forum for European Languages}, pages 222--228. Springer.

\bibitem[{Berant et~al.(2013)Berant, Chou, Frostig, and
  Liang}]{berant2013semantic}
Jonathan Berant, Andrew Chou, Roy Frostig, and Percy Liang. 2013.
\newblock Semantic parsing on {Freebase} from question-answer pairs.
\newblock In \emph{Empirical Methods in Natural Language Processing (EMNLP)}.

\bibitem[{Bromley et~al.(1994)Bromley, Guyon, LeCun, S{\"a}ckinger, and
  Shah}]{bromley1994signature}
Jane Bromley, Isabelle Guyon, Yann LeCun, Eduard S{\"a}ckinger, and Roopak
  Shah. 1994.
\newblock Signature verification using a ``{Siamese}" time delay neural
  network.
\newblock In \emph{NIPS}, pages 737--744.

\bibitem[{Burges et~al.(2005)Burges, Shaked, Renshaw, Lazier, Deeds, Hamilton,
  and Hullender}]{burges2005learning}
Chris Burges, Tal Shaked, Erin Renshaw, Ari Lazier, Matt Deeds, Nicole
  Hamilton, and Greg Hullender. 2005.
\newblock Learning to rank using gradient descent.
\newblock In \emph{Proceedings of the 22nd international conference on Machine
  learning}, pages 89--96.

\bibitem[{Chen et~al.(2017)Chen, Fisch, Weston, and Bordes}]{chen2017reading}
Danqi Chen, Adam Fisch, Jason Weston, and Antoine Bordes. 2017.
\newblock Reading {Wikipedia} to answer open-domain questions.
\newblock In \emph{Association for Computational Linguistics (ACL)}, pages
  1870--1879.

\bibitem[{Das et~al.(2019)Das, Dhuliawala, Zaheer, and McCallum}]{das2019multi}
Rajarshi Das, Shehzaad Dhuliawala, Manzil Zaheer, and Andrew McCallum. 2019.
\newblock Multi-step retriever-reader interaction for scalable open-domain
  question answering.
\newblock In \emph{International Conference on Learning Representations
  (ICLR)}.

\bibitem[{Deerwester et~al.(1990)Deerwester, Dumais, Furnas, Landauer, and
  Harshman}]{deerwester1990indexing}
Scott Deerwester, Susan~T Dumais, George~W Furnas, Thomas~K Landauer, and
  Richard Harshman. 1990.
\newblock Indexing by latent semantic analysis.
\newblock \emph{Journal of the American society for information science},
  41(6):391--407.

\bibitem[{Devlin et~al.(2019)Devlin, Chang, Lee, and
  Toutanova}]{devlin2018bert}
Jacob Devlin, Ming-Wei Chang, Kenton Lee, and Kristina Toutanova. 2019.
\newblock {BERT}: Pre-training of deep bidirectional transformers for language
  understanding.
\newblock In \emph{North American Association for Computational Linguistics
  (NAACL)}.

\bibitem[{Ferrucci(2012)}]{ferrucci2012introduction}
David~A Ferrucci. 2012.
\newblock Introduction to ``{This} is {Watson}".
\newblock \emph{IBM Journal of Research and Development}, 56(3.4):1--1.

\bibitem[{Gillick et~al.(2019)Gillick, Kulkarni, Lansing, Presta, Baldridge,
  Ie, and Garcia-Olano}]{gillick-etal-2019-learning}
Daniel Gillick, Sayali Kulkarni, Larry Lansing, Alessandro Presta, Jason
  Baldridge, Eugene Ie, and Diego Garcia-Olano. 2019.
\newblock Learning dense representations for entity retrieval.
\newblock In \emph{Computational Natural Language Learning (CoNLL)}.

\bibitem[{Guo et~al.(2016)Guo, Kumar, Choromanski, and
  Simcha}]{guo2016quantization}
Ruiqi Guo, Sanjiv Kumar, Krzysztof Choromanski, and David Simcha. 2016.
\newblock Quantization based fast inner product search.
\newblock In \emph{Artificial Intelligence and Statistics}, pages 482--490.

\bibitem[{Guu et~al.(2020)Guu, Lee, Tung, Pasupat, and Chang}]{guu2020realm}
Kelvin Guu, Kenton Lee, Zora Tung, Panupong Pasupat, and Ming-Wei Chang. 2020.
\newblock {REALM}: Retrieval-augmented language model pre-training.
\newblock \emph{ArXiv}, abs/2002.08909.

\bibitem[{Henderson et~al.(2017)Henderson, Al-Rfou, Strope, Sung, Luk{\'a}cs,
  Guo, Kumar, Miklos, and Kurzweil}]{henderson2017efficient}
Matthew Henderson, Rami Al-Rfou, Brian Strope, Yun-hsuan Sung, L{\'a}szl{\'o}
  Luk{\'a}cs, Ruiqi Guo, Sanjiv Kumar, Balint Miklos, and Ray Kurzweil. 2017.
\newblock Efficient natural language response suggestion for smart reply.
\newblock \emph{ArXiv}, abs/1705.00652.

\bibitem[{Huang et~al.(2013)Huang, He, Gao, Deng, Acero, and
  Heck}]{huang2013learning}
Po-Sen Huang, Xiaodong He, Jianfeng Gao, Li~Deng, Alex Acero, and Larry Heck.
  2013.
\newblock Learning deep structured semantic models for {Web} search using
  clickthrough data.
\newblock In \emph{ACM International Conference on Information and Knowledge
  Management (CIKM)}, pages 2333--2338.

\bibitem[{Humeau et~al.(2020)Humeau, Shuster, Lachaux, and
  Weston}]{humeau2020poly}
Samuel Humeau, Kurt Shuster, Marie-Anne Lachaux, and Jason Weston. 2020.
\newblock Poly-encoders: Architectures and pre-training strategies for fast and
  accurate multi-sentence scoring.
\newblock In \emph{International Conference on Learning Representations
  (ICLR)}.

\bibitem[{Izacard and Grave(2020)}]{Izacard2020fid}
Gautier Izacard and Edouard Grave. 2020.
\newblock Leveraging passage retrieval with generative models for open domain
  question answering.
\newblock \emph{ArXiv}, abs/2007.01282.

\bibitem[{Johnson et~al.(2017)Johnson, Douze, and
  J{\'e}gou}]{johnson2017billion}
Jeff Johnson, Matthijs Douze, and Herv{\'e} J{\'e}gou. 2017.
\newblock Billion-scale similarity search with {GPUs}.
\newblock \emph{ArXiv}, abs/1702.08734.

\bibitem[{Joshi et~al.(2017)Joshi, Choi, Weld, and
  Zettlemoyer}]{joshi-etal-2017-triviaqa}
Mandar Joshi, Eunsol Choi, Daniel Weld, and Luke Zettlemoyer. 2017.
\newblock {T}rivia{QA}: A large scale distantly supervised challenge dataset
  for reading comprehension.
\newblock In \emph{Association for Computational Linguistics (ACL)}, pages
  1601--1611.

\bibitem[{Khattab and Zaharia(2020)}]{khattab2020colbert}
Omar Khattab and Matei Zaharia. 2020.
\newblock {ColBERT}: Efficient and effective passage search via contextualized
  late interaction over {BERT}.
\newblock In \emph{ACM SIGIR Conference on Research and Development in
  Information Retrieval (SIGIR)}, pages 39--48.

\bibitem[{Kulis(2013)}]{kulis2013metric}
Brian Kulis. 2013.
\newblock Metric learning: A survey.
\newblock \emph{Foundations and Trends in Machine Learning}, 5(4):287--364.

\bibitem[{Kwiatkowski et~al.(2019)Kwiatkowski, Palomaki, Redfield, Collins,
  Parikh, Alberti, Epstein, Polosukhin, Kelcey, Devlin, Lee, Toutanova, Jones,
  Chang, Dai, Uszkoreit, Le, and Petrov}]{kwiatkowski2019natural}
Tom Kwiatkowski, Jennimaria Palomaki, Olivia Redfield, Michael Collins, Ankur
  Parikh, Chris Alberti, Danielle Epstein, Illia Polosukhin, Matthew Kelcey,
  Jacob Devlin, Kenton Lee, Kristina~N. Toutanova, Llion Jones, Ming-Wei Chang,
  Andrew Dai, Jakob Uszkoreit, Quoc Le, and Slav Petrov. 2019.
\newblock Natural questions: a benchmark for question answering research.
\newblock \emph{Transactions of the Association of Computational Linguistics
  (TACL)}.

\bibitem[{Lee et~al.(2019)Lee, Chang, and Toutanova}]{lee2019latent}
Kenton Lee, Ming-Wei Chang, and Kristina Toutanova. 2019.
\newblock Latent retrieval for weakly supervised open domain question
  answering.
\newblock In \emph{Association for Computational Linguistics (ACL)}, pages
  6086--6096.

\bibitem[{Lewis et~al.(2020{\natexlab{a}})Lewis, Liu, Goyal, Ghazvininejad,
  Mohamed, Levy, Stoyanov, and Zettlemoyer}]{lewis-etal-2020-bart}
Mike Lewis, Yinhan Liu, Naman Goyal, Marjan Ghazvininejad, Abdelrahman Mohamed,
  Omer Levy, Veselin Stoyanov, and Luke Zettlemoyer. 2020{\natexlab{a}}.
\newblock {BART}: Denoising sequence-to-sequence pre-training for natural
  language generation, translation, and comprehension.
\newblock In \emph{Association for Computational Linguistics (ACL)}, pages
  7871--7880.

\bibitem[{Lewis et~al.(2020{\natexlab{b}})Lewis, Perez, Piktus, Petroni,
  Karpukhin, Goyal, K{\"u}ttler, Lewis, Yih, Rockt{\"a}schel, Riedel, and
  Kiela}]{lewis2020rag}
Patrick Lewis, Ethan Perez, Aleksandara Piktus, Fabio Petroni, Vladimir
  Karpukhin, Naman Goyal, Heinrich K{\"u}ttler, Mike Lewis, Wen-tau Yih, Tim
  Rockt{\"a}schel, Sebastian Riedel, and Douwe Kiela. 2020{\natexlab{b}}.
\newblock Retrieval-augmented generation for knowledge-intensive {NLP} tasks.
\newblock In \emph{Advances in Neural Information Processing Systems
  (NeurIPS)}.

\bibitem[{Lin et~al.(2018)Lin, Ji, Liu, and Sun}]{lin2018denoising}
Yankai Lin, Haozhe Ji, Zhiyuan Liu, and Maosong Sun. 2018.
\newblock Denoising distantly supervised open-domain question answering.
\newblock In \emph{Association for Computational Linguistics (ACL)}, pages
  1736--1745.

\bibitem[{Min et~al.(2019{\natexlab{a}})Min, Chen, Hajishirzi, and
  Zettlemoyer}]{min2019discrete}
Sewon Min, Danqi Chen, Hannaneh Hajishirzi, and Luke Zettlemoyer.
  2019{\natexlab{a}}.
\newblock A discrete hard {EM} approach for weakly supervised question
  answering.
\newblock In \emph{Empirical Methods in Natural Language Processing (EMNLP)}.

\bibitem[{Min et~al.(2019{\natexlab{b}})Min, Chen, Zettlemoyer, and
  Hajishirzi}]{min2019knowledge}
Sewon Min, Danqi Chen, Luke Zettlemoyer, and Hannaneh Hajishirzi.
  2019{\natexlab{b}}.
\newblock Knowledge guided text retrieval and reading for open domain question
  answering.
\newblock \emph{ArXiv}, abs/1911.03868.

\bibitem[{Moldovan et~al.(2003)Moldovan, Pa{\c{s}}ca, Harabagiu, and
  Surdeanu}]{moldovan2003performance}
Dan Moldovan, Marius Pa{\c{s}}ca, Sanda Harabagiu, and Mihai Surdeanu. 2003.
\newblock Performance issues and error analysis in an open-domain question
  answering system.
\newblock \emph{ACM Transactions on Information Systems (TOIS)},
  21(2):133--154.

\bibitem[{Mussmann and Ermon(2016)}]{mussmann2016learning}
Stephen Mussmann and Stefano Ermon. 2016.
\newblock Learning and inference via maximum inner product search.
\newblock In \emph{International Conference on Machine Learning (ICML)}, pages
  2587--2596.

\bibitem[{Nie et~al.(2019)Nie, Wang, and Bansal}]{nie2019revealing}
Yixin Nie, Songhe Wang, and Mohit Bansal. 2019.
\newblock Revealing the importance of semantic retrieval for machine reading at
  scale.
\newblock In \emph{Empirical Methods in Natural Language Processing (EMNLP)}.

\bibitem[{Nogueira and Cho(2019)}]{nogueira2019passage}
Rodrigo Nogueira and Kyunghyun Cho. 2019.
\newblock Passage re-ranking with {BERT}.
\newblock \emph{ArXiv}, abs/1901.04085.

\bibitem[{Raffel et~al.(2019)Raffel, Shazeer, Roberts, Lee, Narang, Matena,
  Zhou, Li, and Liu}]{raffel2019T5}
Colin Raffel, Noam Shazeer, Adam Roberts, Katherine Lee, Sharan Narang, Michael
  Matena, Yanqi Zhou, Wei Li, and Peter~J Liu. 2019.
\newblock Exploring the limits of transfer learning with a unified text-to-text
  transformer.
\newblock \emph{ArXiv}, abs/1910.10683.

\bibitem[{Rajpurkar et~al.(2016)Rajpurkar, Zhang, Lopyrev, and
  Liang}]{rajpurkar2016squad}
Pranav Rajpurkar, Jian Zhang, Konstantin Lopyrev, and Percy Liang. 2016.
\newblock {SQuAD}: 100,000+ questions for machine comprehension of text.
\newblock In \emph{Empirical Methods in Natural Language Processing (EMNLP)},
  pages 2383--2392.

\bibitem[{Ram and Gray(2012)}]{ram2012maximum}
Parikshit Ram and Alexander~G Gray. 2012.
\newblock Maximum inner-product search using cone trees.
\newblock In \emph{Proceedings of the 18th ACM SIGKDD international conference
  on Knowledge discovery and data mining}, pages 931--939.

\bibitem[{Roberts et~al.(2020)Roberts, Raffel, and Shazeer}]{roberts2020much}
Adam Roberts, Colin Raffel, and Noam Shazeer. 2020.
\newblock How much knowledge can you pack into the parameters of a language
  model?
\newblock \emph{ArXiv}, abs/2002.08910.

\bibitem[{Robertson and Zaragoza(2009)}]{robertson2009probabilistic}
Stephen Robertson and Hugo Zaragoza. 2009.
\newblock The probabilistic relevance framework: {BM25} and beyond.
\newblock \emph{Foundations and Trends in Information Retrieval},
  3(4):333--389.

\bibitem[{Seo et~al.(2019)Seo, Lee, Kwiatkowski, Parikh, Farhadi, and
  Hajishirzi}]{seo2019real}
Minjoon Seo, Jinhyuk Lee, Tom Kwiatkowski, Ankur Parikh, Ali Farhadi, and
  Hannaneh Hajishirzi. 2019.
\newblock Real-time open-domain question answering with dense-sparse phrase
  index.
\newblock In \emph{Association for Computational Linguistics (ACL)}.

\bibitem[{Shrivastava and Li(2014)}]{NIPS2014_5329}
Anshumali Shrivastava and Ping Li. 2014.
\newblock Asymmetric {LSH} ({ALSH}) for sublinear time maximum inner product
  search ({MIPS}).
\newblock In \emph{Advances in Neural Information Processing Systems (NIPS)},
  pages 2321--2329.

\bibitem[{Voorhees(1999)}]{voorhees1999trec}
Ellen~M Voorhees. 1999.
\newblock The {TREC-8} question answering track report.
\newblock In \emph{TREC}, volume~99, pages 77--82.

\bibitem[{Wang et~al.(2018)Wang, Yu, Guo, Wang, Klinger, Zhang, Chang, Tesauro,
  Zhou, and Jiang}]{wang2018r}
Shuohang Wang, Mo~Yu, Xiaoxiao Guo, Zhiguo Wang, Tim Klinger, Wei Zhang, Shiyu
  Chang, Gerry Tesauro, Bowen Zhou, and Jing Jiang. 2018.
\newblock R\^{}3: Reinforced ranker-reader for open-domain question answering.
\newblock In \emph{Conference on Artificial Intelligence (AAAI)}.

\bibitem[{Wang et~al.(2019)Wang, Ng, Ma, Nallapati, and Xiang}]{wang2019multi}
Zhiguo Wang, Patrick Ng, Xiaofei Ma, Ramesh Nallapati, and Bing Xiang. 2019.
\newblock Multi-passage {BERT}: A globally normalized bert model for
  open-domain question answering.
\newblock In \emph{Empirical Methods in Natural Language Processing (EMNLP)}.

\bibitem[{Wolfson et~al.(2020)Wolfson, Geva, Gupta, Gardner, Goldberg, Deutch,
  and Berant}]{wolfson2020break}
Tomer Wolfson, Mor Geva, Ankit Gupta, Matt Gardner, Yoav Goldberg, Daniel
  Deutch, and Jonathan Berant. 2020.
\newblock Break it down: A question understanding benchmark.
\newblock \emph{Transactions of the Association of Computational Linguistics
  (TACL)}.

\bibitem[{Xiong et~al.(2020{\natexlab{a}})Xiong, Xiong, Li, Tang, Liu, Bennett,
  Ahmed, and Overwijk}]{Xiong2020ANCE}
Lee Xiong, Chenyan Xiong, Ye~Li, Kwok-Fung Tang, Jialin Liu, Paul Bennett,
  Junaid Ahmed, and Arnold Overwijk. 2020{\natexlab{a}}.
\newblock Approximate nearest neighbor negative contrastive learning for dense
  text retrieval.
\newblock \emph{ArXiv}, abs/2007.00808.

\bibitem[{Xiong et~al.(2020{\natexlab{b}})Xiong, Wang, and
  Wang}]{Xiong2020ProgressivelyPD}
Wenhan Xiong, Hankang Wang, and William~Yang Wang. 2020{\natexlab{b}}.
\newblock Progressively pretrained dense corpus index for open-domain question
  answering.
\newblock \emph{ArXiv}, abs/2005.00038.

\bibitem[{Yang et~al.(2019{\natexlab{a}})Yang, Xie, Lin, Li, Tan, Xiong, Li,
  and Lin}]{yang2019end}
Wei Yang, Yuqing Xie, Aileen Lin, Xingyu Li, Luchen Tan, Kun Xiong, Ming Li,
  and Jimmy Lin. 2019{\natexlab{a}}.
\newblock End-to-end open-domain question answering with bertserini.
\newblock In \emph{North American Association for Computational Linguistics
  (NAACL)}, pages 72--77.

\bibitem[{Yang et~al.(2019{\natexlab{b}})Yang, Xie, Tan, Xiong, Li, and
  Lin}]{Yang2019Data}
Wei Yang, Yuqing Xie, Luchen Tan, Kun Xiong, Ming Li, and Jimmy Lin.
  2019{\natexlab{b}}.
\newblock Data augmentation for bert fine-tuning in open-domain question
  answering.
\newblock \emph{ArXiv}, abs/1904.06652.

\bibitem[{Yih et~al.(2011)Yih, Toutanova, Platt, and Meek}]{yih2011learning}
Wen-tau Yih, Kristina Toutanova, John~C Platt, and Christopher Meek. 2011.
\newblock Learning discriminative projections for text similarity measures.
\newblock In \emph{Computational Natural Language Learning (CoNLL)}, pages
  247--256.

\end{thebibliography}
\bibliographystyle{acl_natbib}
\clearpage %
\appendix

\setcounter{table}{4} %

\section{Distant Supervision}
\label{sec:distant}

When training our final \model/ model using Natural Questions, we use the passages in our collection that best match the gold context as the positive passages.
As some QA datasets contain only the question and answer pairs, it is thus interesting to see when using the passages that contain the answers as positives (i.e., the distant supervision setting), whether there is a significant performance degradation.
Using the question and answer together as the query, we run Lucene-BM25 and pick the top passage that contains the answer as the positive passage.
Table~\ref{tab:qa_ir_dist_sv} shows the performance of \model/ when trained using the original setting and the distant supervision setting.

\section{Alternative Similarity Functions \& Triplet Loss}
\label{sec:alt-sim}
\label{sec:triplet}

In addition to dot product (DP) and negative log-likelihood based on softmax (NLL), we also experiment with Euclidean distance (L2) and the triplet loss.
We negate L2 similarity scores before applying softmax and change signs of question-to-positive and question-to-negative similarities when applying the triplet loss on dot product scores. 
The margin value of the triplet loss is set to 1.
Table~\ref{tab:qa_ir_sim_and_loss} summarizes the results.
All these additional experiments are conducted using the same hyper-parameters tuned for the baseline (DP, NLL).

Note that the retrieval accuracy for our ``baseline" settings reported in Table~\ref{tab:qa_ir_dist_sv} (Gold) and Table~\ref{tab:qa_ir_sim_and_loss} (DP, NLL) is slightly better than those reported in Table 3.
This is due to a better hyper-parameter setting used in these analysis experiments, which is documented in our code release.

\section{Qualitative Analysis}
\label{sec:qual_ana}

Although \model/ performs better than BM25 in general, the retrieved passages of these two retrievers actually differ qualitatively. Methods like BM25 are sensitive to highly selective keywords and phrases, but cannot capture lexical variations or semantic relationships well.  In contrast, \model/ excels at semantic representation, but might lack sufficient capacity to represent salient phrases which appear rarely.  Table~\ref{tab:retrieved-examples} illustrates this phenomenon with two examples.  In the first example, the top scoring passage from BM25 is irrelevant, even though keywords such as \textit{England} and \textit{Ireland} appear multiple times.  In comparison, \model/ is able to return the correct answer, presumably by matching \textit{``body of water"} with semantic neighbors such as \textit{sea} and \textit{channel}, even though no lexical overlap exists.
The second example is one where BM25 does better.  The salient phrase \textit{``Thoros of Myr"} is critical, and \model/ is unable to capture it.

\begin{table}[t!]
    \setlength\tabcolsep{5pt}
    \centering
    \begin{tabular}{lcccc} \toprule
         & \tf{Top-1} & \tf{Top-5} & \tf{Top-20} & \tf{Top-100}  \\ \midrule
        {Gold} & 44.9 & 66.8 & 78.1 & 85.0 \\
        {Dist. Sup.} & 43.9 & 65.3 & 77.1 & 84.4 \\
    \bottomrule
    \end{tabular}
    \caption{
    Retrieval accuracy on the development set of Natural Questions, trained on passages that match the gold context (Gold) or the top BM25 passage that contains the answer (Dist. Sup.).
    }
    \label{tab:qa_ir_dist_sv}
\end{table}

\begin{table}[t!]
    \setlength\tabcolsep{4pt}
    \centering
    \begin{tabular}{ll|cccc}
    \toprule
    \tf{Sim} & \tf{Loss} & \multicolumn{4}{c}{\tf{Retrieval Accuracy}} \\ 
    & & Top-1 & Top-5 & Top-20 & Top-100 \\ \midrule
    \multirow{2}{*}{DP} & NLL & \tf{44.9} & \tf{66.8} & \tf{78.1} & \tf{85.0} \\
    & Triplet & 41.6 & 65.0 & 77.2 & 84.5 \\
    \midrule
    \multirow{2}{*}{L2} & NLL & 43.5 & 64.7 & 76.1 & 83.1 \\
    & Triplet & 42.2 & 66.0 & \tf{78.1} & 84.9 \\ 
    \bottomrule
    \end{tabular}
     \caption{Retrieval Top-$k$ accuracy on the development set of Natural Questions using different similarity and loss functions.} 

    \label{tab:qa_ir_sim_and_loss}
\end{table}
\setlength{\tabcolsep}{6pt} %

\begin{table*}[!t]
\scriptsize
\begin{tabular}{p{3cm}|p{6cm}|p{6cm}}
\toprule
\tf{Question} & \tf{Passage received by BM25} & \tf{Passage retrieved by \model/} \\
\midrule
\multirow{2}{3cm}{What is the body of water between England and Ireland?} & Title:British Cycling  & Title: Irish Sea\\
& \ldots  \tf{England} is not recognised as a region by the UCI, and there is no English cycling team outside the Commonwealth Games. For those occasions, British Cycling selects and supports the \tf{England} team. Cycling is represented on the Isle of Man by the Isle of Man Cycling Association.  Cycling in Northern \tf{Ireland} is organised under Cycling Ulster, part of the all-Ireland governing \tf{body} Cycling \tf{Ireland}. Until 2006, a rival governing \tf{body} existed,  \ldots & \ldots Annual traffic between Great Britain and \tf{Ireland} amounts to over 12 million passengers and of traded goods. {\color{blue}\tf{The Irish Sea}} is connected to the North Atlantic at both its northern and southern ends. To the north, the connection is through the North Channel between Scotland and Northern \tf{Ireland} and the Malin Sea. The southern end is linked to the Atlantic through the St George's Channel between Ireland and Pembrokeshire, and the Celtic Sea. \ldots \\
\midrule
\multirow{2}{3cm}{Who plays Thoros of Myr in Game of Thrones?} & Title: No One (Game of Thrones) & Title:
P\r{a}l Sverre Hagen \\ & \ldots He may be "no one," but there's still enough of a person left in him to respect, and admire who this girl is and what she's become. Arya finally tells us something that we've kind of known all along, that she's not no one, she's Arya Stark of Winterfell." "No One" saw the reintroduction of Richard Dormer and {\color{blue}\tf{Paul Kaye}}, who portrayed Beric Dondarrion and \tf{Thoros} of \tf{Myr}, respectively, in the third season, \ldots &
P\r{a}l Sverre Valheim Hagen (born 6 November 1980) is a Norwegian stage and screen actor. He appeared in the Norwegian film "Max Manus" and played Thor Heyerdahl in the Oscar-nominated 2012 film "Kon-Tiki". Pål Hagen was born in Stavanger, Norway, the son of Roar Hagen, a Norwegian cartoonist who has long been associated with Norway\'s largest daily, "VG". He lived in Jåtten, a neighborhood in the city of Stavanger in south-western Norway. \ldots \\
\bottomrule
\end{tabular}
\caption{Examples of passages returned from BM25 and \model/. Correct answers are written in {\color{blue} \tf{blue}} and the content words in the question are written in bold.}
\label{tab:retrieved-examples}
\end{table*}

\section{Joint Training of Retriever and Reader}
\label{sec:joint}

We fix the passage encoder in our joint-training scheme while allowing only the question encoder to receive backpropagation signal from the combined (retriever + reader) loss function.
This allows us to leverage the HNSW-based FAISS index for efficient low-latency retrieving, without reindexing the passages during model updates.
Our loss function largely follows ORQA's approach, which uses log probabilities of positive passages selected from the retriever model, and correct spans and passages selected from the reader model.
Since the passage encoder is fixed, we could use larger amount of retrieved passages when calculating the retriever loss.
Specifically, we get top 100 passages for each question in a mini-batch and use the method similar to in-batch negative training: all retrieved passages' vectors participate in the loss calculation for \emph{all} questions in a batch. 
Our training batch size is set to 16, which effectively gives 1,600 passages per question to calculate retriever loss.
The reader still uses 24 passages per question, which are selected from the top 5 positive and top 30 negative passages (from the set of top 100 passages retrieved from the same question).
The question encoder's initial state is taken from a \model/ model previously trained on the NQ dataset.
The reader's initial state is a BERT-base model.
In terms of the end-to-end QA results, our joint-training scheme does not provide better results compared to the usual retriever/reader training pipeline, resulting in the same 39.8 exact match score on NQ dev as in our regular reader model training.

\end{document}